\documentclass[runningheads]{llncs}
\usepackage{graphicx}
\usepackage{amsmath,amssymb} 
\usepackage{color}
\usepackage{epsfig}
\usepackage{graphicx}
\usepackage{array}
\usepackage{booktabs}
\usepackage{xcolor}
\usepackage{cite}
\usepackage{multirow}

\begin{document}
\title{
    Joint Person Segmentation and Identification in \\
    Synchronized First- and Third-person Videos
} 

\titlerunning{Person Segmentation and Identification in First- and Third-person Videos}
%
\author{Mingze Xu, Chenyou Fan, Yuchen Wang, Michael S. Ryoo, David J. Crandall}
\institute{
  School of Informatics, Computing, and Engineering \\
    Indiana University, Bloomington, IN 47408 \\
    \email{\{mx6, fan6, wang617, mryoo, djcran\}@indiana.edu}
}
%
\authorrunning{Xu \textit{et al}.}
%
\maketitle              
\newcommand{\fp}{first-person\ }
\newcommand{\tp}{third-person\ }
\newcommand{\fandtp}{first- and third-person\ }
\newcommand{\etal}{\textit{et al}.}
\newcommand{\xhdr}[1]{\noindent {\textbf{#1}}}
\newcommand{\ra}[1]{\renewcommand{\arraystretch}{#1}}

\begin{abstract}
    In a world of pervasive cameras, public spaces are often captured
    from multiple perspectives by cameras of different types, both fixed
    and mobile. An important problem is to organize these
    heterogeneous collections of videos by finding connections between
    them, such as identifying correspondences between the people
    appearing in the videos and the people holding or wearing the
    cameras. In this paper, we wish to solve two specific problems:
    (1) given two or more synchronized third-person
    videos of a scene, produce a pixel-level segmentation of each
    visible person and identify corresponding people across different
    views (i.e., determine who in camera A corresponds with whom in
    camera B), and (2) given one or more synchronized third-person
    videos as well as a first-person video taken by a mobile or wearable camera,
    segment and identify the camera wearer in the third-person videos.
    Unlike previous work which requires ground truth bounding boxes to
    estimate the correspondences, we perform person
    segmentation and identification jointly. We find that solving these two
    problems simultaneously is mutually beneficial, because better
    fine-grained segmentation allows us to better perform matching
    across views, and information from multiple views helps us
    perform more accurate segmentation. We evaluate our approach on two
    challenging datasets of interacting people captured from multiple
    wearable cameras, and show that our proposed method performs
    significantly better than the state-of-the-art on both person
    segmentation and identification.
\keywords{Synchronized first- and third-person cameras.}
\end{abstract}
\section{Introduction}

There will be an estimated 45 \textit{billion} cameras on Earth by
2022 --- more than five times the number of people~\cite{ldv}! In a
world with so many cameras, it will be commonplace for a scene to be
simultaneously recorded by multiple cameras of different types. For
example, a busy city street may be recorded by not only fixed
surveillance cameras, but also by mobile cameras on smartphones,
laptops, tablets, self-driving cars, and even wearable devices like
GoPro~\cite{gopro} and Snap Spectacles~\cite{snap}. As cameras
continue to multiply, new techniques will be needed to 
organize and make sense of these weakly-structured collections of video.
For example, a key problem in many applications is to detect,
identify, and track people. Combining data from
multiple cameras could significantly improve performance on this and
other scene understanding problems, since evidence from multiple
viewpoints could help resolve ambiguities caused by occlusion,
perspective distortion, etc. However, integrating evidence across
heterogeneous cameras in unconstrained dynamic
environments is a challenge, especially for wearable and
mobile devices where the camera is moving unpredictably.

\begin{figure}[t]
    \center
    \includegraphics[width=0.95\columnwidth,trim=0cm 1.5cm 0cm 0.5cm,clip]{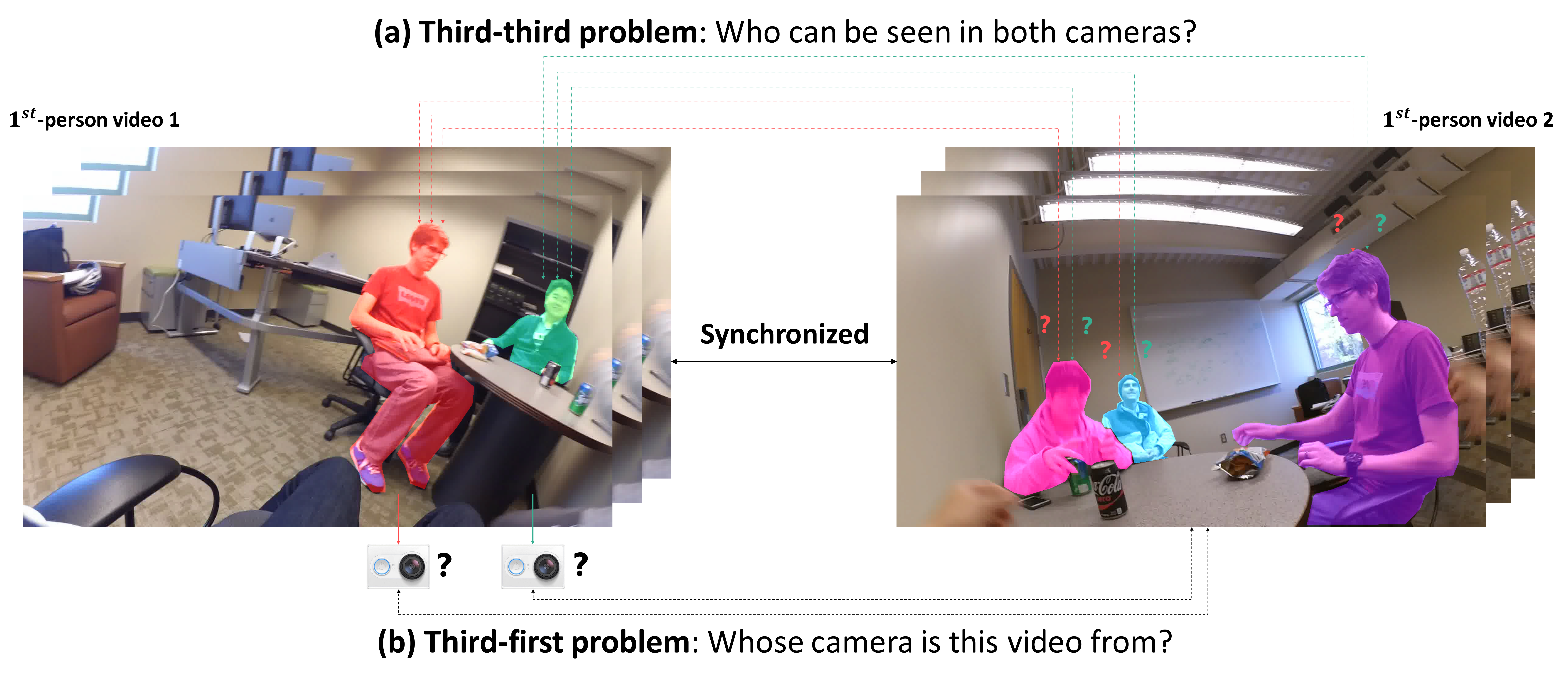}
    \caption{
        Two or more people move around an environment while wearing cameras. We
        are interested in two specific, related problems: (a) given one or more
        synchronized third-person videos of a scene, segment all the visible
        people and identify corresponding people across the different videos;
        and (b) given one or more synchronized third-person videos of a scene
        as well as a video that was taken by a wearable first-person camera,
        identify and segment the person who was wearing the camera in the
        third-person videos.
    }
    \label{fig:ad}
\end{figure}

For example, consider a law enforcement scenario in which multiple police
officers chase a suspect through a crowded square. Body-worn police
cameras (which nearly 95\% of U.S. police departments use or plan to
deploy~\cite{bodywornpolice}) record events from the officers'
perspectives. Investigators later want to reconstruct the incident by combining
the first-person wearable camera videos with third-person views
from surveillance cameras and civilian smartphone videos uploaded to
social media. In any given frame of any given camera, they
may want to identify: (1) fine-grained, pixel-level segmentation masks for all
people of interest, including both the suspect and the officers
(e.g., for activity or action recognition), (2) the
instances in which one of the camera wearers (officers) was visible in
another camera's view, and (3) instances of the same person appearing in
different views at the same time. The scene is complex and crowded, requiring
fine-grained segmentation masks to separate individual people (since frequent
occlusions would cause  bounding boxes to overlap). The wearable
camera videos are particularly challenging because the cameras themselves are moving
rapidly.

While person tracking and (re-)identification are well-studied in computer
vision~\cite{persontracking,reidentification}, only  recently have they
been considered in challenging scenarios of heterogeneous
first-person  and traditional cameras.
Ardeshir and Borji~\cite{ucf-eccv16} consider the case of several people moving
around  while wearing cameras, and try to match each of these
first-person views to one of the people appearing in a third-person, overhead
view of the scene. This is challenging because the camera wearer is
never seen in their own wearable video, so he or she must be identified by
matching their motion from a third-person perspective with the first-person
visual changes that are induced by their movements. That paper's approach is
applicable in closed settings with overhead cameras  (e.g., a
 museum), but not in unconstrained environments such as our law
enforcement example. Fan \etal~\cite{firstthird2017cvpr}
relax many assumptions, allowing
arbitrary third-person camera views and including evidence based on scene
appearance. Zheng \etal~\cite{zheng2017iccv} consider the distinct 
 problem of identifying the same person appearing in multiple wearable
camera videos (but not trying to identify the camera wearers themselves). But
these techniques identify individual people using bounding boxes, which
are too coarse in crowded scenes with frequent occlusions.
Moreover, these techniques
assume that accurate oracle bounding boxes are available (even at test time).

In this paper, we consider the more challenging problem of
not only finding correspondences between people in first- and
third-person cameras, but also producing pixel-level segmentation
masks of the people in each view (see Fig.~\ref{fig:ad}). We
define a \textit{first-person} camera to be a wearable camera for which we
care about the identity of the camera wearer, while a \textit{third-person}
camera is either a static \textit{or} wearable camera for which we are
\textit{not} interested in determining the wearer. Our
hypothesis is that  simultaneous segmentation and matching
is mutually beneficial: segmentation helps refine matching by
producing finer-grained appearance features (compared to 
bounding boxes), which are important in crowded scenes with
many occlusions, while matching helps
locate a person of interest and produce better segmentation masks,
which in turn help in tasks like activity and action
recognition. We show that previous work~\cite{firstthird2017cvpr}
is a special case of ours, since we can naturally
handle their \fandtp cases. We evaluate on two publicly
available datasets augmented with pixel-level annotations,
showing that we achieve significantly better results than
 numerous baselines.

\section{Related Work}
We are not aware of work on joint person segmentation and
identification in  \fandtp cameras, so we draw
inspiration from several related problems.

\textit{Object Segmentation in Images and Videos.}
Deep learning has achieved state-of-the-art performance on semantic image
segmentation~\cite{long2015fully,badrinarayanan2015segnet,zhao2016pyramid,lin2017refinenet,chen2018encoder},
typically using fully convolutional networks (FCNs) that extract low-resolution
features and then up-sample.
Other approaches~\cite{pinheiro2015learning,pinheiro2016learning,li2016fully,he2017mask}
are based on region proposals, inspired by  R-CNNs~\cite{girshick2015fast,ren2015faster}
for object detection. For example, Mask R-CNNs~\cite{he2017mask} 
separately predict object masks and their class
labels, avoiding competition among classes and improving performance for
overlapped instances. For 
object
segmentation in video~\cite{khoreva2016learning,voigtlaender2017online,caelles2017one,Jang_2017_CVPR,Koh_2017_CVPR,tokmakov2017learning},
most methods assume that the object
mask in the first frame is known (during both training and testing) and the task is to
propagate them to subsequent frames. Khoreva \etal~\cite{perazzi2017learning} propose 
guided instance segmentation that uses the object mask from the
previous frame to predict the next one. The network is pre-trained (off-line)
on static images and fine-tuned (on-line) on the first frame's annotations 
for specific objects of interest. 
We follow a similar formulation, except that 
we incorporate both  appearance and optical flow in a two-stream
network,
helping to better update the object mask across time. 
Our work is also inspired by the pixel-level Siamese matching network of
Yoon \etal~\cite{yoon2017pixel}
that segments and identifies objects, even 
those not seen during training. 
We extend to multiple cameras by using object instances across
multiple synchronized videos to learn
variations and correspondences in appearance across views.
Cheng
\etal~\cite{Cheng_ICCV_2017} propose a two-stream network which outputs 
segmentation and optical flow simultaneously, where segmentation
focuses on objectness and optical flow  exploits motion. Inspired by their observation that segmentation and optical flow
benefit each other, we propose a novel architecture that jointly performs
person segmentation and identification.

\textit{Co-segmentation.}  Our work is related to co-segmentation of
objects appearing in multiple images~\cite{rother2006cosegmentation} or
videos~\cite{rubio2012video,chen2012video,chiu2013multi,fu2014object,guo2014consistent}. Several
methods use Markov Random Fields with a regularized difference of
feature histograms, for example, by assuming a Gaussian prior on the
objectness appearance~\cite{rother2006cosegmentation} or computing 
sum squared differences~\cite{batra2010icoseg}.  
Chiu \etal~\cite{chiu2013multi} use
distance-dependent Chinese Restaurant Processes as priors on both
appearance and motion for unsupervised (not semantic)
co-segmentation. Fu \etal~\cite{fu2014object} address video
co-segmentation as CRF inference on an object co-selection
graph, but segmentation candidates are computed only by a
category-independent method~\cite{endres2010category}
and are not refined from information across multiple videos. Guo
\etal~\cite{guo2014consistent} perform iterative constrained
clustering using seed superpixels and pairwise constraints, and refine
the segmentation with a multi-class MRF. Most of
these methods assume that either a target object appears in all videos
or that videos contain at least one common target object, and none
apply deep learning.  To the best of our knowledge, ours is
the first paper to propose a deep learning approach to co-segmentation
in videos, and is applicable both to single and multiple camera
scenarios.

\textit{First-person Cameras.}
Ardeshir and Borji~\cite{ucf-eccv16} match a set of \fp
videos to a set of people appearing in a top-view video using
graph matching, but assume there are multiple \fp cameras
sharing the same field of view at any time and only consider
\tp cameras that are overhead. 
 Fan \etal~\cite{firstthird2017cvpr}
identify a  \fp camera wearer in a \tp video using a
two-stream semi-Siamese network that incorporates spatial and temporal
information from both views, and learns a joint embedding
space from 
\fandtp matches. Zheng \etal~\cite{zheng2017iccv} identify people
appearing in multiple wearable camera videos (but do not
identify the camera wearers themselves). 

The above work assumes that the people  have been
detected with accurate bounding boxes in both training \textit{and
test} datasets. We build on these methods, proposing a
novel architecture that simultaneously segments and identifies 
camera wearers and others. We find that segmenting and identifying
are mutually beneficial; in the law scenario
described above with crowded scenes and 
occluded people, for example, fine-grained segmentation masks are needed to
accurately extract visual features specific to any given 
person, while identity information from multiple views helps 
accurately segment the person in any individual view.

\section{Our Approach}
Given two or more videos taken from a set of cameras (potentially 
both static and wearable cameras), we wish to segment each person
appearing in these videos, identify matches between segments that correspond
to the same person across different views, and identify the segments that
correspond to the wearer of each \fp camera. The main idea is that
despite having very different perspectives, synchronized cameras recording the
same environment should be capturing some of the same people and background objects.
This overlap  permits finding similarities and correspondences among these
videos in both visual and motion domains, as long as differences
caused by differing  viewpoints are ignored. Unlike prior work~\cite{firstthird2017cvpr} which
assumes a ground truth bounding box is available for each person in each
frame, we perform segmentation and matching simultaneously. We
hypothesize that these two tasks are mutually beneficial: 
person segmentations provide more accurate information than
coarse bounding boxes for people matching, while 
people's appearance and motion from different perspectives 
produce better segmentation masks.

More concretely, we formulate our problem as two separate
tasks. The \textit{third-third problem} is to segment each person and
find person correspondences across different views captured from a
pair of \tp cameras.  The \textit{third-first problem} is to segment
and identify the camera wearer of a given \fp video in \tp videos.  We
first introduce a basic network architecture for both 
problems: a two-stream fully convolutional network (FCN) that
estimates a segmentation mask for each person using the current RGB
frame, stacked optical flow fields, and segmentation result of
the previous frame (which we call the \textit{pre-mask})
(Section~\ref{sec:fcn}). We then introduce a Siamese network for each of our two problems,
that incorporates the FCN and allows
person segmentation and identification to benefit each other
(Section~\ref{sec:siamese}). Finally we describe our
loss used for segmentation and distance
metric learning (Section~\ref{sec:loss}).

\begin{figure}[t]
    \center
    \includegraphics[width=0.95\columnwidth,clip,trim=0cm 0cm 0cm 0.5cm]{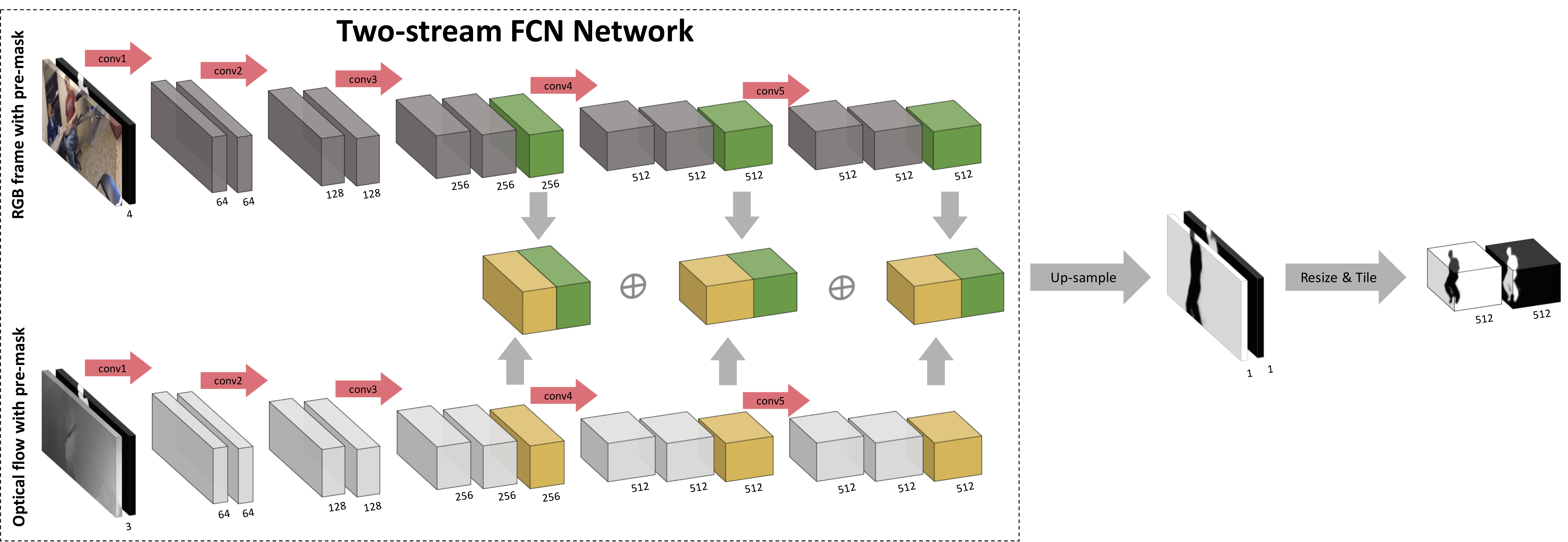}
    \caption{
        Visualization of our two-stream FCN network.
        We feed RGB frames with pre-masks to the visual stream (top, dark
        grey) and stacked optical flow fields with pre-mask to
        the motion stream (bottom, light grey). The spatial and temporal
        features at \texttt{pool3}, \texttt{pool4}, and \texttt{pool5} are
        fused to predict the segmentation of the target person. We
        downsample the extracted features of the \texttt{softmax} layer by 
        16, then tile the background and foreground channels by 512, separately.
    }
    \label{fig:fcn}
\end{figure}

\subsection{Two-stream FCN Network}
\label{sec:fcn}
We use FCN8s~\cite{long2015fully} as the basis of our framework but
with several important modifications.  We chose FCN8s due to their
effectiveness and compactness, although other architectures such as
DeepLabv3+~\cite{chen2018encoder} and Mask R-CNN~\cite{he2017mask}
could be easily used.  Fig.~\ref{fig:fcn} presents our novel
architecture.  To take advantage of video and incorporate evidence
from both appearance and motion, we expand FCN8s to a two-stream
architecture, where a \textit{visual stream} receives RGB frames (top
of Fig.~\ref{fig:fcn}) and a \textit{motion stream} receives stacked
optical flow fields (bottom).  This design is inspired by Simonyan and
Zisserman~\cite{simonyan2014two}, although their network was proposed
for a completely different problem (action recognition from a single
static camera).  To jointly consider both spatial and temporal
information, we use ``early'' fusion to concatenate features at levels
\texttt{pool3}, \texttt{pool4}, and \texttt{pool5} (middle of
Fig.~\ref{fig:fcn}).  Following FCN8s to incorporate ``coarse, high
level information with fine, low level
information''~\cite{long2015fully} for more accurate segmentation, we
combine the fused features from these different levels.

However, in contrast to Long \textit{et al.}'s FCN8s, our
two-stream FCN  targets instance segmentation: we want to segment specific people, not just all
instances of the ``{person}'' class.  We address this with
an instance-by-instance strategy in both training and test,
in which we only consider a single person at a time. In
order to guide the network to segment a specific person among the many
that may appear in a frame, we append that person's binary
pre-mask (without any semantic information) to the input of each
stream as an additional channel.  This pre-mask
provides a rough estimate of the person's location and his or
her approximate shape in the current frame.
In training,
our network is pre-trained by taking ground truth pre-masks as
inputs, and then fine-tuned with estimated masks from the previous frame.
In testing, we assume that we have a (possibly quite coarse) segmentation 
of each person  in the first frame 
and propagate this mask forward by evaluating 
each subsequent unlabeled frame in sequence. 
 A pixel-level classification loss function is used
to guide learning  
(Section~\ref{sec:loss}). 

\subsection{Siamese Networks}
\label{sec:siamese}

The network in the last section learns to estimate the segmentation
mask of a specific person across frames of video.  We now
use this network in a Siamese structure with a contrastive
loss to match person instances across different third- and
first-person views.  The main idea behind our Siamese networks
is to learn an embedding space such that features
captured by different cameras from different perspectives are close
together only if they actually belong to the same person --- i.e.,
so that a person's
appearance features are invariant to camera viewpoint.
The Siamese formulation allows us
to simultaneously learn the viewpoint-invariant embedding space 
for matching identities and the pixel-wise segmentation network
described above in an end-to-end fashion.
Moreover, our Siamese (or semi-Siamese) FCN architecture improves the
invariance of object segmentation across different perspectives and
transformations. In contrast to co-segmentation methods that require
pairs of images or videos in both training and testing, our approach
only need pairs in the training phase.  In
testing, our two-stream FCN network can be applied to any single
stream input, and uses the embedding space to match with others. To
allow the segmentation network to receive arbitrary sizes of inputs, our
contrastive loss function is generalized to a 3D representation
space, with a Euclidean distance for
positive exemplars and a hinge loss for negative ones.

\begin{figure}[t]
    \center
    \includegraphics[width=0.95\columnwidth,clip,trim=0cm 0cm 0cm 0cm]{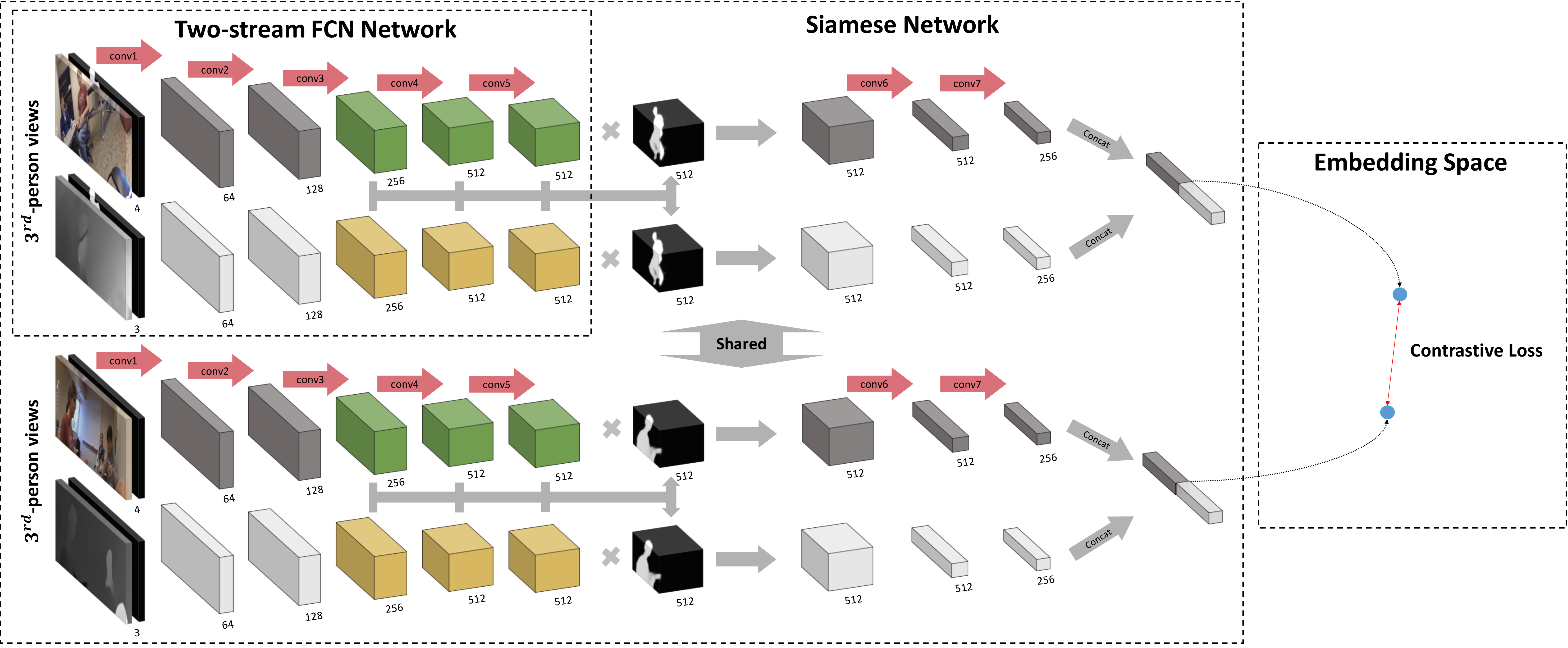}
    \caption{
        Our third-third network segments and identifies
        the people in common across different videos. The network is composed
        of two FCN branches with a Siamese structure, where all convolution
        layers (shown in the same color) are shared.
    }
    \label{fig:third-third}
\end{figure}

In particular, we explore two  Siamese network structures,
customized for our two tasks: the third-third problem of segmenting
and matching people across a pair of cameras, and the
third-first problem of segmenting a person of interest and identifying
if he or she is the wearer of a \fp camera. The third-third problem
considers a more general case in which the cameras may be static or
may be wearable, but they are all viewing a person of interest from a
\tp viewpoint; we thus use a full-Siamese network that shares all
convolution layers in the FCN branch and the embedding layers.
In contrast, the third-first problem
must match feature representations from different perspectives
(identifying how a camera wearer's ego-motion visible in a \fp view
correlates with that same motion's appearance from a \tp
view). As in~\cite{firstthird2017cvpr}, our third-first network
is formulated in a semi-Siamese structure, where separate shallow
layers capture different low-level features while
deeper ones are shared. 

\textit{Third-third Network.}
Fig.~\ref{fig:third-third} shows the architecture of our third-third network,
which segments and matches people in common from a pair of \tp
camera views. We use a Siamese structure with two branches of the FCN network from
Fig.~\ref{fig:fcn} (and discussed in Section~\ref{sec:fcn}), where all corresponding
convolution layers are shared. The Siamese branch is thus encouraged to learn
relationships between people's appearance in different views by optimizing a generalized embedding space. The key
idea is that despite being captured from very different perspectives, the same
person in synchronized videos should have some correspondences in both visual
and motion domains.

\begin{figure}[t]
    \center
    \includegraphics[width=0.95\columnwidth,clip,trim=0cm 0cm 0cm 0cm]{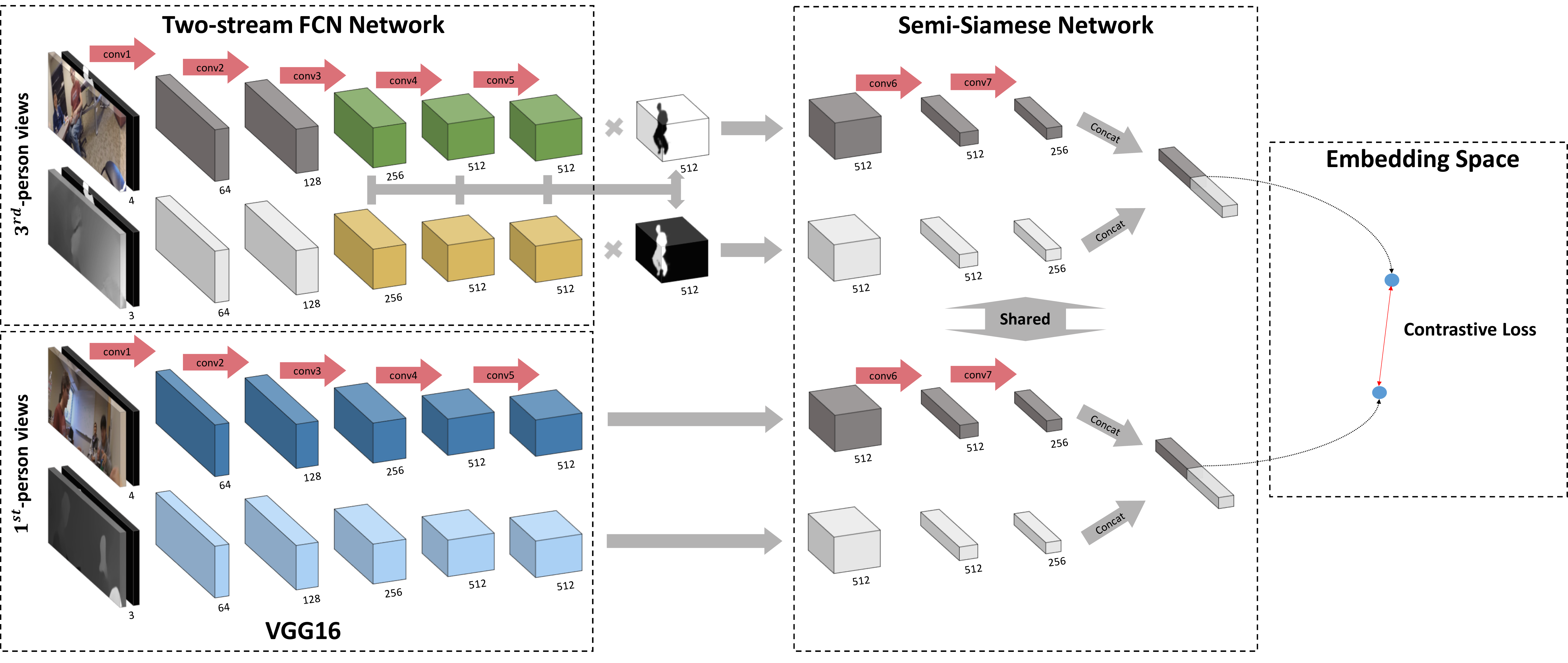}
    \caption{
        Our third-first network segments and identifies the \fp camera wearer
        in \tp videos. The network is formulated in a
        semi-Siamese structure where only convolution layers of the embedding
        space (shown in the same color) are shared.
    }
    \label{fig:third-first}
\end{figure}

In more detail, given an RGB frame and optical flow fields (appended
with the pre-mask of the person of interest) as inputs, each of size
$W \times H$, the FCN branch estimates a binary-valued person
segmentation mask of the same size.  The Siamese branch is then
appended to the \texttt{pool5} layer of both visual and motion streams
with an input size of $512 \times W' \times H'$, where
$W'=\frac{W}{16}$ and $H'=\frac{H}{16}$, for matching. To obtain more
accurate representations for each ``target'' person, we re-weight the
spatial and temporal features by multiplying them with the confidence
outputs of the FCN branch. To emphasize the pixel positions belonging
to the person while retaining some contextual information, we use soft
attention maps after the softmax layer rather than the estimated
segmentation mask. We first resize the soft attention of the
foreground from $1 \times W \times H$ to $1 \times W' \times H'$ and
tile it to $512 \times W' \times H'$ to fit the size of \texttt{pool5}
outputs.  For both visual and motion streams, we multiply this resized
confidence map with the features, which gives a higher score to the person's 
pixels and a low score to the background. 
By ``cropping out'' the region corresponding to a
person from the feature maps, the match across two views should
receive a higher correspondence. This
correspondence will also back-propagate its confidence to improve
segmentation.  Finally, the re-weighted spatial and
temporal features are concatenated together for matching each person instance.

\textit{Third-first Network.}
Fig.~\ref{fig:third-first} shows the architecture of our third-first
network, the goal of which is to segment a \fp camera wearer in \tp videos
and to recognize the correspondence between the \fp view and its
representation in \tp videos. To be specific, given a \fp video, our
network must decide which, if anyone, of the people appearing in a \tp video is
the wearer of this \fp camera, and to estimate the wearer's
segmentation. In contrast to the third-third network which has two
FCN branches focusing on the same task (person segmentation), the
second branch of the third-first network receives the \fp videos
as inputs and is designed to extract the wearer's ego-motion and the
visual information of the background, which hopefully also provides
constraints for the segmentation. We thus propose a semi-Siamese
network to learn the \fandtp distance metric, where the \fp branch has
a similar structure to the FCN but without the up-sampling layers or
the segmentation loss. The structure of the Siamese branch is similar
to that of the third-third network, but with a different re-weighting
method: we multiply the spatial features with the soft attention of
the background but the temporal features with the soft attention of
the foreground. We do this because   camera wearers
do not appear in their own \fp videos (with occasional exceptions
of arms or hands), but the backgrounds reflect some similarities
between different perspectives; meanwhile, motion features of camera
wearers in \tp videos is related to the ego-motion 
in \fp videos. The re-weighted
appearance and motion features are then concatenated after several
convolution operations, as
discussed above.

\subsection{Loss Functions}
\label{sec:loss}
We propose two loss functions for joint segmentation and distance metric optimization
for a batch of $N$ training exemplars.
First, \textit{sigmoid cross entropy loss}
compares a predicted segmentation
mask to ground truth,
\begin{equation}
    \begin{split}
        L_{seg} = - \sum^N_i \sum^W_w \sum^H_h \big( S_{i,w,h} \cdot \log
        \hat{S}_{i,w,h} + (1-S_{i,w,h}) \cdot \log (1-\hat{S}_{i,w,h}) \big),
    \end{split}
\end{equation}
where $\hat{S_i} \in \{0,1\}^{W \times H}$ is the
predicted segmentation mask of
exemplar $i$ and
$S_i \in
\{0,1\}^{W \times H}$ is the corresponding ground truth mask.
Second, \textit{generalized contrastive loss} encourages low distances between 
positive exemplars (pairs of corresponding people)
and high distances between negative ones,
\begin{equation}
    \begin{split}
        L_{siam} = \sum^N_i\sum^C_c\sum^{W''}_w\sum^{H''}_h \, &y_i \, ||a_{i,c,w,h}-b_{i,c,w,h}||^2 + \, \\
                                                               & (1-y_i) \max(m-||a_{i,c,w,h}-b_{i,c,w,h}||, \, 0)^2,
    \end{split}
\end{equation}
where $m$ is a constant, $a_i$ and $b_i$ are two features
corresponding to exemplar $i$, and  $y_i$ is
$1$ if $i$ is a correct correspondence and $0$ otherwise.
This loss enables our model to learn an embedding space for arbitrary
input sizes.

\section{Experiments}

We test our third-third and third-first networks on joint person
segmentation and identification in two datasets of 
synchronized \fandtp videos, collected by two different authors.
We primarily evaluate on the publicly available IU ShareView dataset~\cite{firstthird2017cvpr}, consisting of 9 sets of
two  5-10 minute \fp videos. Each set contains 3-4
participants performing a variety of everyday activities 
(shaking hands, chatting, eating, etc.) in one of six indoor
environments. Each person in each frame is annotated with a ground truth
bounding box and a unique person ID. To evaluate our methods on person
segmentation, we manually augmented a subset of the dataset with 
pixel-level person annotations, for a total of 1,277 labeled frames containing 2,654
annotated person instances. We computed optical flow
fields for all videos using  FlowNet2.0~\cite{ilg2017flownet}.

Since adjacent frames are typically highly correlated, we split the training
and test datasets at the video level, with 6 video sets used for training (875
annotated frames) and 3 sets for testing (402 annotated frames). In
each set of videos, there are 3-4 participants, two of which wear \fp
cameras. Note that a \fp camera never sees its own wearer, so the
people not wearing cameras are the ones who are in common across the \fp
videos. Since our approach uses sequences of contiguous frames and pairs of
instances (either a pair of two people or a pair of one person and one
camera view), we divide each video set into several short sequences, each
with 10-15 consecutive frames. More specifically, in training we create 484
positive and 1,452 negative pairs for the third-third problem, and 865 positive
and 1,241 negative pairs for the third-first problem (about a 1:3 ratio).
In testing, each problem has 10 sequences of pairs of videos, and each
video has 20 consecutive frames (about 4 seconds).
Thus we have about 400 annotated test frames for evaluating matching, and about
1,000 person instances for evaluating segmentation (since every frame has 2-3
people).

We also evaluate our models on a subset of UTokyo Ego-Surf~\cite{yonetani2015ego},
which contains 8 diverse groups of \fp videos recorded synchronously during
face-to-face conversations in both indoor and outdoor environments. Limited by
the size of the dataset (only 3 available pairs of short videos including 3-4
participants), we use it only for testing, and still train on IU ShareView.
As before, 
we manually created pixel-level person annotations for 10 sequences of pairs of
videos, each with 20 consecutive frames.

\subsection{Evaluation Protocol}
\label{sec:settings}

We implemented our networks in PyTorch~\cite{pytorch}, and performed
all experiments on a single Nvidia Titan X Pascal GPU.

\textit{Training.} Our training process consisted of two stages: (a) optimizing only the FCN
branch supervised by the pixel-level classifier for
providing imperfect but reasonable soft attentions, and (b) optimizing the
joint model (either the third-third or third-first network) based on the person
segmentation and identification tasks, simultaneously. Our two-stream FCN
network is built on VGG16~\cite{simonyan2014very}, and 
we initialized both visual and motion streams
using weights pre-trained on ImageNet~\cite{imagenet_cvpr09}. The
FCN branch was then optimized with an instance-by-instance strategy,
which only considers one particular person of interest at a time, and
uses the ground truth pre-mask as an additional channel to indicate
which person the network should focus on. We used stochastic gradient
descent (SGD)  with fixed learning rate $10^{-4}$, momentum
$0.9$, weight decay $0.0005$, and batch size $25$. Learning was
terminated after $30$ epochs. Our joint model was
then initialized with the weights of the pre-trained FCN and 
fine-tuned by considering pairs of instances as inputs for person
segmentation and identification. We again used SGD optimization
but with learning rate $10^{-5}$. For the first $20$
epochs, we froze the weights of the FCN branch, and optimized the
Siamese branch to make the contrastive loss converge to a
``reasonable'' range (not too large to destroy the soft attention). We
then started the joint learning process, and terminated after
\textit{another} $40$ epochs.

\textit{Testing.}
In contrast to training, which requires pairs of videos as
inputs, our joint model can be applied to an individual stream, where each video
frame is processed to simultaneously estimate each person's segmentation and
extract corresponding features for matching between different streams.
In testing, all possible pairs of instances are considered as candidate matches: each pair
contains either two people from different videos in the third-third problem, or
a \fp camera view and a person appearing in a \tp video in the third-first problem.
Unlike methods that require a pair of instances as input, our approach
only needs to process each person and camera view once.

\begin{figure}[t]
    \includegraphics[width=1\columnwidth,trim=0cm 0cm 0cm 0cm,clip]{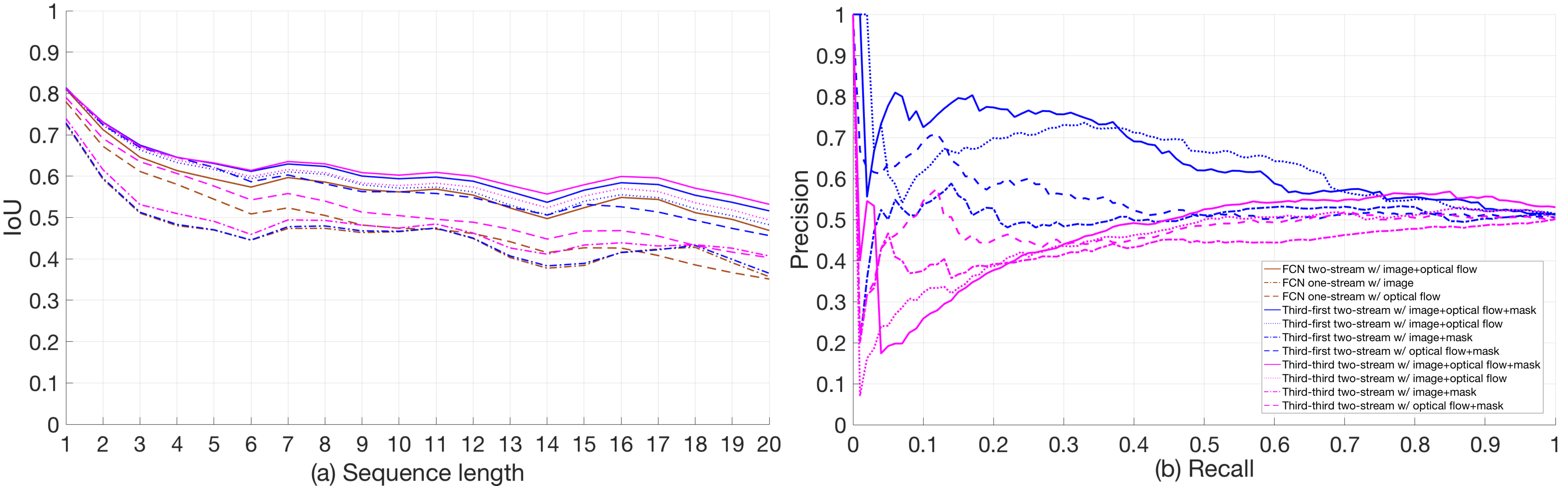}
    \caption{
        IoU and precision-recall curves of our models on
        IU ShareView dataset~\cite{firstthird2017cvpr}
    }
    \label{fig:plot}
\end{figure}

\subsection{Evaluation}
For both third-first and third-third problems, we evaluate our method with two
tasks: person (1) segmentation and (2) identification across multiple cameras.

\textit{Person Segmentation}
is evaluated
in terms of \textit{intersection over union (IoU)} between the estimated
segmentation maps and the ground truth. This is measured over each video in
the test dataset by applying our models to each frame. Our model
sequentially takes the segmentation results from the previous frame (the pre-mask)
as input to guide the segmentation of the next frame. In the evaluation, the ground segmentation
mask of the first (and only the first) video frame is assumed to be available.

\textit{Person Identification} is evaluated
with \textit{Mean average precision (mAP)} and \textit{Accuracy (ACC)}, each of
which takes a different view of the problem. mAP treats
people matching as a retrieval problem: given all possible pairs
of person instances from two different cameras (i.e., two person instances from
different \tp videos in the third-third problem \textit{or} one person instance
from \tp video and one  \fp video in the third-first problem), we wish
to retrieve all pairs corresponding to the same person.
ACC evaluates whether the single best match
for a given candidate is correct:  for every person instance in each view,
the classifier is forced to choose a single matching instance in all other views,
and we calculate the percentage of matches that are correct.
This setting is the same to the one used in Fan \etal~\cite{firstthird2017cvpr}, except that 
their task is significantly easier because they assume person ground-truth
bounding boxes are available during both training and testing, whereas
our approach must infer the person's position (as well as segmentation mask) automatically.

\begin{table}[t]
    \caption{
        Experimental results of our models on IU ShareView dataset~\cite{firstthird2017cvpr}}
        {\scriptsize{\textsf{
        \begin{tabular}{@{}lcccccccccccccc@{}} \toprule      
            \multicolumn{8}{c}{\textbf{Network Architecture}} & &  \multicolumn{5}{c}{\textbf{Evaluation}} \\
            \cmidrule{1-8}
            \cmidrule{10-14}
            \multicolumn{3}{c}{\textit{}} & &\multicolumn{2}{c}{\textit{Streams}} & & & &\multicolumn{1}{c}{\textit{Segmentation}} & & \multicolumn{3}{c}{\textit{Identification}} \\
            \cmidrule{5-6}
            \cmidrule{10-10}
            \cmidrule{12-14}
            & \quad\quad & Backbone & & \, Image \, & \,Optical flow\, & \quad\quad & Re-weighting & \quad\quad\quad & IoU & \quad\quad & mAP & & ACC \\ \midrule
            \multirow{4}{*}{Baselines}
            & & Copy First &    & &                    & & - & & 41.9 & & - & & - \\
            & & FCN && X &               & & - & & 47.1 & & - & & - \\
            & & FCN &&   & X       & & - & & 50.9 & & - & & - \\
            & & FCN && X & X & & - & & 57.3 & & - & & - \\ \midrule
            \multirow{5}{*}{Third-Third}
            & & VGG && X & X  & & bounding box~\cite{firstthird2017cvpr} & & - & & 44.2 & & 40.1 \\
            & & FCN && X &               & & soft attention & & 49.3 & & 44.3 & & 44.5 \\
            & & FCN &&   & X       & & soft attention & & 54.1 & & 48.4 & & 46.2 \\
            & & FCN && X & X & & w/o  & & 60.6 & & 45.6 & & 48.9 \\
            & & FCN && X & X & & soft attention & & \textbf{62.7} & & \textbf{49.0} & & \textbf{55.5} \\ \midrule
            \multirow{5}{*}{Third-First}
            & & VGG && X & X & & bounding box~\cite{firstthird2017cvpr} & & - & & 64.1 & & 50.6 \\
            & & FCN && X &      & & soft attention & & 47.4  & & 51.4 & & 52.7 \\
            & & FCN &&   & X        & & soft attention & & 58.9  & & 55.1 & & 53.1 \\
            & & FCN && X & X & & w/o  & & 59.8 & & 64.0 & & 61.7 \\
            & & FCN && X & X & & soft attention & & \textbf{61.9} & & \textbf{65.2} & & \textbf{73.1} \\
            \bottomrule
        \end{tabular}
}}}
    \label{tab:exp}
\end{table}

\begin{table}[t]
    \caption{
        Experimental results of our models on UTokyo Ego-Surf dataset~\cite{yonetani2015ego}
    }
    {\scriptsize{\textsf{
        \begin{tabular}{@{}lcccccccccccccc@{}} \toprule      
            \multicolumn{8}{c}{\textbf{Network Architecture}} & &  \multicolumn{5}{c}{\textbf{Evaluation}} \\
            \cmidrule{1-8}
            \cmidrule{10-14}
            \multicolumn{3}{c}{\textit{}} & &\multicolumn{2}{c}{\textit{Streams}} & & &  &\multicolumn{1}{c}{\textit{Segmentation}} & & \multicolumn{3}{c}{\textit{Identification}} \\
            \cmidrule{5-6}
            \cmidrule{10-10}
            \cmidrule{12-14}
            & \quad\quad & Backbone & & \,\,\,\, Image \,\,\,\, & \,\,\,\,Optical flow\,\,\,\, & \quad\quad & Re-weighting & \quad\quad\quad & IoU & \quad\quad & mAP & & ACC \\ \midrule
            \multirow{2}{*}{Third-Third}
            & & FCN && X & X & & w/o  & & 42.1 & & 43.8 & & 36.7 \\
            & & FCN && X & X & & soft attention & & \textbf{43.0} & & \textbf{45.5} & & \textbf{42.0} \\ \midrule
            \multirow{2}{*}{Third-First}
            & & FCN && X & X & & w/o  & & 41.4 & & 45.2 & & 44.0 \\
            & & FCN && X & X & & soft attention & & \textbf{43.6} & & \textbf{52.0} & & \textbf{55.2} \\
            \bottomrule
        \end{tabular}
}}}
    \label{tab:extra}
\end{table}

\subsection{Experimental Results}

\textit{Baselines.}
To characterize the difficulty of segmentation in this dataset, we first test several baselines, shown in
Table~\ref{tab:exp} for IU ShareView.
{Copy First} simply propagates the
ground truth segmentation mask from the first frame to all following frames in
the sequence. In a completely static scenes with no motion, the IoU
of Copy First should be 100.0, but our dataset includes
frequent motion of both the wearable cameras and people,
and thus shows a relatively low IoU of 41.9.
A second baseline consisting of a single-stream FCN using only image information
achieves somewhat better IoU of 47.1,
while a third baseline consisting of a single-stream FCN using only optical flow achieves 50.9. A two-stream
baseline FCN that combines both visual and motion performs significantly better than either
one-stream network, achieving IoU of 57.3.

\textit{Our Models.}
We next test our approach that jointly performs
segmentation with person instance matching. On segmentation,
our full model produces an IoU of 62.7 for the third-third
scenario and 61.9 for third-first, compared to 57.3 for the
two-stream baseline that performs only segmentation.
Fig.~\ref{fig:plot} (a) reports more detailed analysis of the segmentation
performance (Y-axis) based on the length of video sequences (X-axis), and shows
that our approach is still able to predict reasonable results on long videos. 
To permit a fair comparison
across models, both the one- and two-stream FCNs were optimized with the
same hyper-parameters (discussed in Section~\ref{sec:settings}).
Table~\ref{tab:exp} also presents results on person instance
matching on IU ShareView.  We achieved mAP scores of 49.0 and 65.2 on the third-third
and third-first problems, respectively, and ACCs of 55.5 and
73.1. We compare these results with the state-of-the-art
method of Fan \etal~\cite{firstthird2017cvpr}.  Their task is
to match \fp camera views to camera wearers in \textit{static} \tp
video, so we extend it to our third-third and third-first problems by  re-implementing their best
model using VGG16~\cite{simonyan2014very} (instead of
AlexNet~\cite{krizhevsky2012imagenet}) and training on our new,
augmented dataset. The results show that our joint model outperforms
in both third-third (mAP of 49.0 vs. 44.2) and third-first (mAP of 65.2 vs. 64.1)
problems. This is likely due to learning a more accurate embedding
space, with the help of jointly learning to perform segmentation. More importantly, our
approach is able to obtain more accurate feature representations from
people's pixel-level locations rather than simply relying on rough
bounding boxes.  
Fig.~\ref{fig:plot} (b) compares the
precision-recall curves of the different techniques for person matching.

\textit{UTokyo Ego-Surf Dataset.}  We also test our models on our subset
of UTokyo Ego-Surf (without retraining), and
Table~\ref{tab:extra} summarizes the results. Though performing worse
than on the IU ShareView dataset on which they were trained, the
models still give reasonable results, indicating robustness even
though the datasets are recorded by different cameras (Xiaoyi Yi
vs. Panasonic HX-A500) and scenarios (indoor vs. outdoor).

\textit{Ablation Studies.}
We also test simpler variants of our technique.
To evaluate our re-weighting method that incorporates estimated soft
attention maps, we tried \textit{not} re-weighting the spatial and temporal
features and simply using \texttt{pool5} layer outputs. We also compare with
the results of~\cite{firstthird2017cvpr}, which uses ground truth bounding
boxes to ``crop out'' regions of interest. As shown in
Table~\ref{tab:exp} and~\ref{tab:extra}, using re-weighting with
soft attention not only outperforms for the matching task but also 
generates better segmentation maps.
Our ablation study also tested the relative contribution of each of our motion and visual
feature streams. As shown in Table~\ref{tab:exp}, our dual-stream approach
performs significantly better than either single-stream optical flow or visual information
on both the third-third and third-first problems, evaluated on both segmentation and
matching.

\begin{figure}[t]
    \center
    \includegraphics[width=0.95\columnwidth]{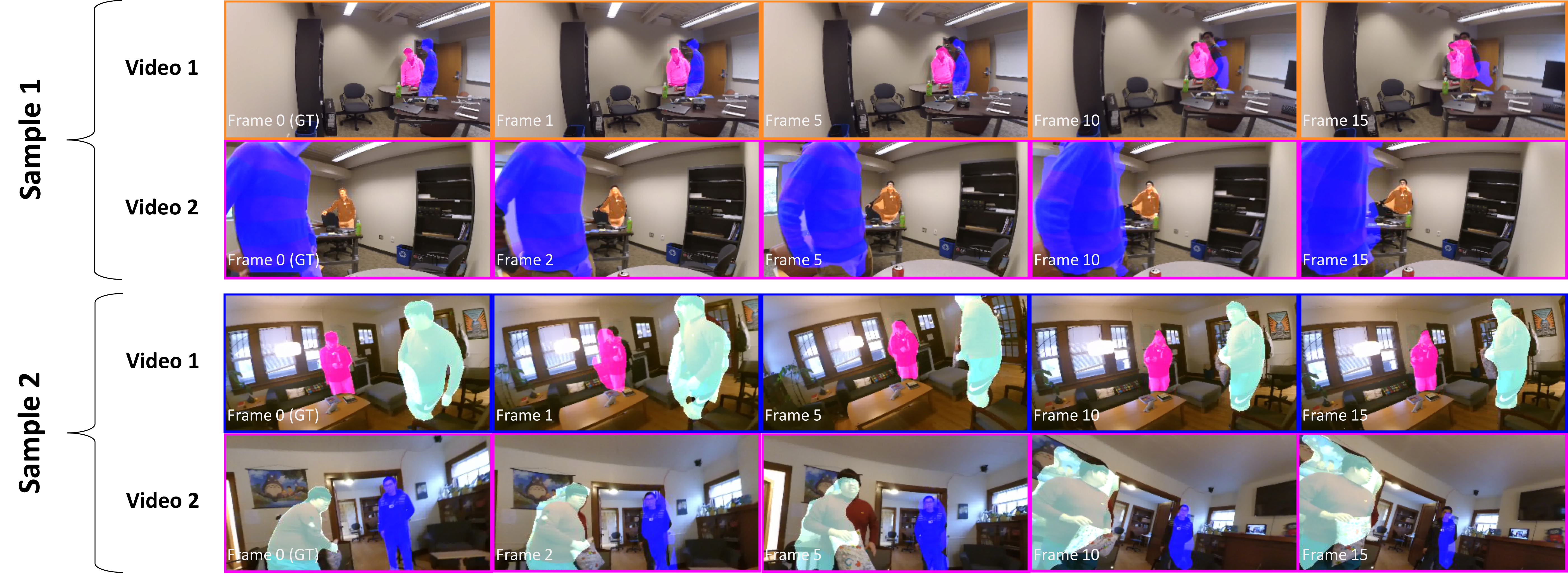}
    \caption{
        Sample results of the third-third and third-first problems, where
        two videos of each sample are from two synchronized wearable
        cameras. The color of person segmentation masks and camera views
        indicates the correspondences across different cameras.
    }
    \label{fig:sample_results}
\end{figure}

\section{Conclusion}
We presented a novel fully convolutional network (FCN) with Siamese
and semi-Siamese structures for joint person instance segmentation and
identification. 
We also prepared a new, challenging dataset with
person pixel-level annotations and correspondences in multiple first- and
third-person cameras. Our results
demonstrated the effectiveness and robustness of our approach on
joint  person segmentation and identification. The
results suggested that jointly inferring pixel-level segmentation maps
and correspondences of people helps perform each individual task more
accurately, and that incorporating both visual and motion information
works better than either individually.

Although our results are encouraging, our techniques have limitations
and raise opportunities for future work.  First, the joint models
assume people appear in every frame of the video, so that our approach
will treat someone who disappears from the scene and then re-enters as
a new person instance. While this assumption is reasonable for the
relatively short video sequences we consider here, future work could
easily add a re-identification module to recognize people who have
appeared in previous frames. Second, the joint models perform a FCN
forward pass for every individual person in each frame; future work
could explore sharing computation costs to improve the efficiency of
our method, especially for real-time applications.  Lastly, we plan to
further evaluate our approach on larger datasets including more
diverse scenarios.

\subsubsection{Acknowledgments.}
This work was supported by the National Science Foundation (CAREER IIS-1253549), and
the IU Office of the Vice Provost for Research, the College of Arts
and Sciences, and the School of Informatics, Computing, and
Engineering through the Emerging Areas of Research Project ``Learning:
Brains, Machines, and Children.''  
We would like to thank Sven Bambach for assisting with dataset
collection, and Katherine Spoon and Anthony Tai for suggestions on our paper draft.

%
%
%
%
%
%
%
%
\bibliographystyle{splncs04}
\bibliography{egbib}
\end{document}